\documentclass[letterpaper,conference]{ieeeconf} 

\IEEEoverridecommandlockouts                              

\usepackage{lipsum}  
\usepackage{lastpage} 
\usepackage{fancyhdr} 
\usepackage{hyperref}
\usepackage[letterpaper, top=1cm, bottom=1cm, left=1.5cm, right=1.5cm, headheight=15pt, headsep=10pt, includehead, includefoot]{geometry}

\usepackage{graphicx}
\usepackage{algorithm}
\usepackage{algpseudocode}
\usepackage{amsmath, amssymb, mathtools}
\usepackage{cancel}
\usepackage[most]{tcolorbox}
\usepackage{xcolor}
\usepackage{booktabs} 
\usepackage{multirow} 
\usepackage[caption=false, font=footnotesize]{subfig}
\usepackage[normalem]{ulem}
\usepackage{flushend}

\definecolor{codegray}{rgb}{0.5,0.5,0.5}
\definecolor{colorbackgroud}{rgb}{1,1,1}
\definecolor{borderblue}{rgb}{0.36,0.36,0.56}
\definecolor{codegreen}{rgb}{0.26,0.76,0.26}
\definecolor{codeblue}{rgb}{0.9, 0.9, 0.9}
\definecolor{longhri}{rgb}{0.6, 0.6, 0.6}
\definecolor{dmpcolor}{rgb}{0.75, 0.75, 0.75}

\newtcolorbox{codebox}{
    before = \vspace{5pt},
    colback=colorbackgroud, 
    colframe=borderblue, 
    boxrule=0.5pt, 
    arc=0pt, 
    left=0pt, 
    right=0pt, 
    boxsep=0pt, 
    top=1pt,           
    bottom=1pt,
    breakable,
    enhanced jigsaw,
}
\newtcolorbox{codebox1}{
    before = \vspace{5pt},
    colback=colorbackgroud, 
    colframe=borderblue, 
    boxrule=0.5pt, 
    arc=0pt, 
    left=0pt, 
    right=0pt, 
    boxsep=0pt, 
    top=1pt,           
    bottom=1pt,
    breakable,
    enhanced jigsaw,
}

\begin{document}
\bstctlcite{IEEEexample:BSTcontrol}
\title{\textbf{Enhancing the LLM-Based Robot Manipulation \\ Through Human-Robot Collaboration}}

\author{Haokun~Liu,
    Yaonan~Zhu$^{\ast}$, Kenji Kato, Atsushi Tsukahara, Izumi Kondo, \\ Tadayoshi Aoyama, and Yasuhisa~Hasegawa
\thanks{This work was supported in part by JST SICORP, Japan, under Grant JPMJSC2305; and in part by JSPS KAKENHI under Grant JP24K17236; and in part by NCGG under Chojuiryou Kenkyukaihatsuhi Nos. 19–5, 21-21. \textit{($^{\ast}$Corresponding author: Yaonan Zhu)}}
\thanks{Haokun Liu is with Department of Mechanical Systems Engineering, Nagoya University, Nagoya 464-8603, Japan (e-mail: haokun@robo.mein.nagoya-u.ac.jp).}
\thanks{Yaonan Zhu is with the School of Engineering, The University of Tokyo, Tokyo 113-8656, Japan, and Department of Micro-Nano Mechanical Science and Engineering, Nagoya University, Nagoya 464-8603, Japan (yaonan.zhu@weblab.t.u-tokyo.ac.jp).}
\thanks{Tadayoshi Aoyama and Yasuhisa Hasegawa are with Department of Micro-Nano Mechanical Science and Engineering, Nagoya University, Nagoya 464-8603, Japan (e-mail: \{aoyama, hasegawa\}@mein.nagoya-u.ac.jp).}
\thanks{Kenji Kato, Atsushi Tsukahara and Izumi Kondo are with National Center for Geriatrics and Gerontology, Obu, Aichi, 474-8511, Japan (e-mail: \{kk0724, tsukahara, ik7710\}@ncgg.go.jp).}
\thanks{Digital Object Identifier (DOI): https://doi.org/10.1109/LRA.2024.3415931}
}

\markboth{IEEE ROBOTICS AND AUTOMATION LETTERS. PREPRINT VERSION. ACCEPTED May, 2024}%
{Liu \MakeLowercase{\textit{et al.}}: Enhancing the LLM-Based Robot Manipulation through Human-Robot Collaboration} 


\maketitle
\thispagestyle{fancy}
\begin{abstract}
Large Language Models (LLMs) are gaining popularity in the field of robotics. 
However, LLM-based robots are limited to simple, repetitive motions due to the poor integration between language models, robots, and the environment. 
This paper proposes a novel approach to enhance the performance of LLM-based autonomous manipulation through Human-Robot Collaboration (HRC). 
The approach involves using a prompted GPT-4 language model to decompose high-level language commands into sequences of motions that can be executed by the robot. 
The system also employs a YOLO-based perception algorithm, providing visual cues to the LLM, which aids in planning feasible motions within the specific environment.
Additionally, an HRC method is proposed by combining teleoperation and Dynamic Movement Primitives (DMP), allowing the LLM-based robot to learn from human guidance. 
Real-world experiments have been conducted using the Toyota Human Support Robot for manipulation tasks. The outcomes indicate that tasks requiring complex trajectory planning and reasoning over environments can be efficiently accomplished through the incorporation of human demonstrations.
\end{abstract}


\section{Introduction}
%
%
%
%
The concept of autonomous robots capable of interacting with humans via natural language has captivated the field of robotics research for years. 
Recently, the emergence of Large Language Models (LLMs), powered by Transformer technology, has presented a practical approach to realize this vision \cite{vaswani2023attention}.
Over the years, LLMs have emerged as a key technology in machine learning, demonstrating remarkable capabilities in language processing \cite{touvron2023llama}. Notably, LLMs have shown considerable promise in areas such as human-robot interaction and task planning \cite{saycan2022arxiv}, \cite{wu2023tidybot}. 

For LLMs to effectively guide robot behavior in real-world scenarios, integrating environmental perception is crucial. This synergy enables the robot to autonomously interpret and act upon user instructions in real-world scenarios. Key technologies that facilitate this include computer vision algorithms like You Only Look Once (YOLO) \cite{redmon2016look}, and innovative approaches such as Contrastive Language-Image Pre-Training (CLIP) \cite{radford2021learning} and the Segment Anything Model (SAM) \cite{kirillov2023segment}. The combination of LLMs with advanced computer vision has spurred the development of sophisticated robotic manipulation platforms capable of executing complex, precise actions based on user commands given by natural languages.
 \begin{figure}[t]
    \centering
    \includegraphics[width=0.87\linewidth]{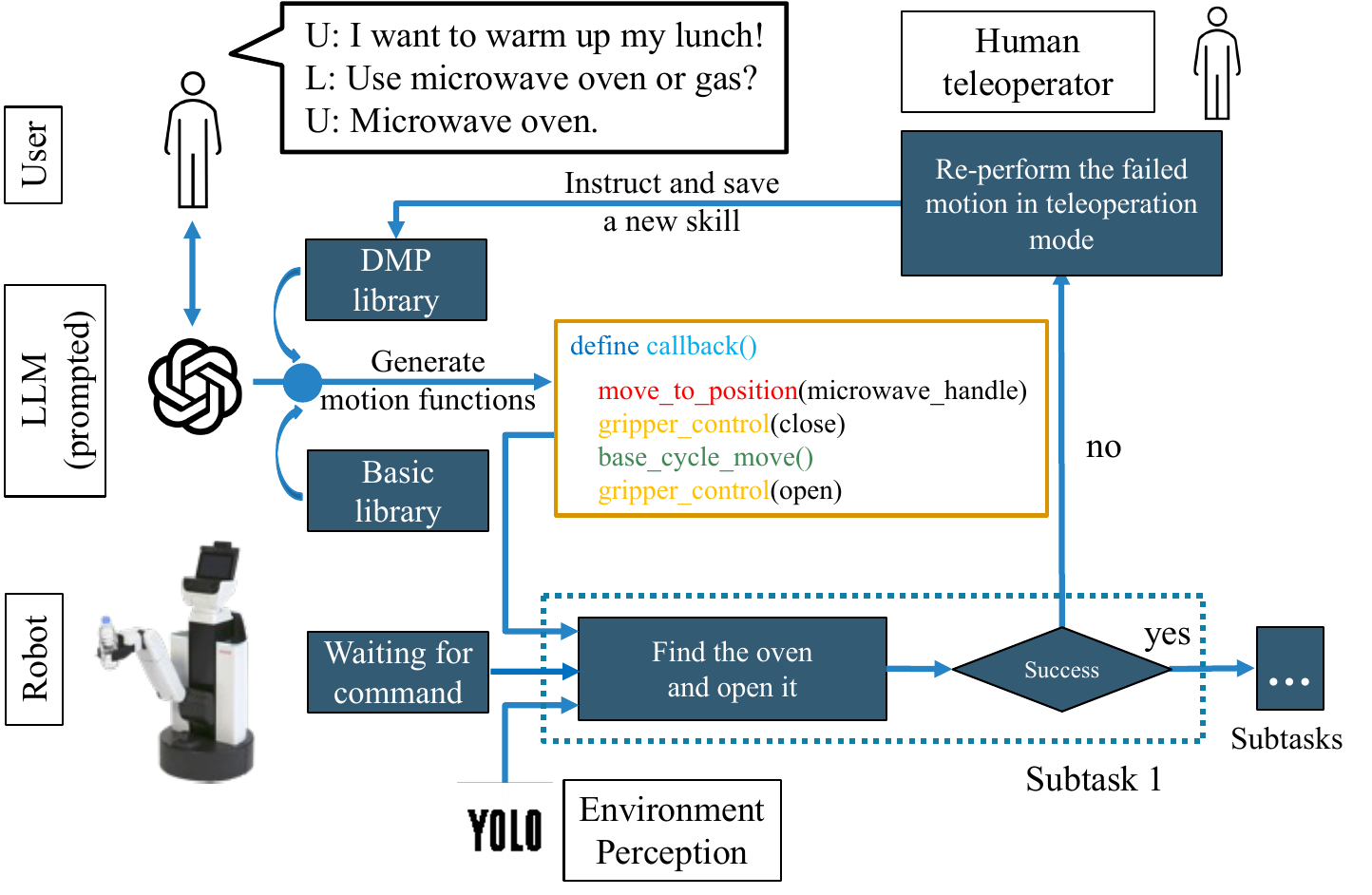}
    \captionsetup{skip=0pt}
    \caption{An overview of an LLM-based Human-Robot Collaboration System, featuring user interaction, a basic library for pre-programmed motion functions, and a DMP library for adaptive motion function generation and storage to accomplish a complex real-world task. (e.g. ``warm up my lunch'')}
    \label{fig:system}
 \vspace{-6mm}
\end{figure}


Despite the impressive natural language-based interaction and logical reasoning capabilities of LLMs, existing research in the real-world LLM-based robot systems primarily focuses on basic and straightforward task planning such as picking and placing \cite{singh2023progprompt,zhao2023differentiable}, the more intricate, long-horizon tasks, such as warming and roasting food, which require high-level reasoning and motion planning, have often been overlooked. 
These shortcomings are primarily due to the following challenges:
(i) The prevalent method for LLM-based robot control is through generating codes that trigger robots' motion. Undoubtedly, this approach is a straightforward way to accomplish robot manipulation via LLM. However, this approach falters when dealing with complex trajectories, as it lacks the advanced capability to intelligently modify motion commands to meet specific, dynamic environment requirements \cite{singh2023progprompt}. This limitation is a significant hurdle in enhancing robotic autonomy and responsiveness.
(ii) In current LLM-based human-robot interaction frameworks, human commands are typically restricted to single-instance inputs, limiting the scope for human continuous supervision and instructions. This constraint hinders the potential of these systems for complex, dynamic real-world tasks. Thus, intuitive and seamless Human-Robot Collaboration (HRC) is essential to extend the channels of input and enhance the capabilities of LLM-based robots.

To address the challenges aforementioned, this paper proposes a novel LLM-based robot manipulation framework combined with HRC, as illustrated in Fig. 1. 
This system advances beyond simple LLM-based autonomy by integrating a teleoperation system. This feature allows user input during the autonomous process, ensuring human guidance for complex tasks. Through Dynamic Movement Primitives (DMP), the system captures and stores trajectory data from manual teleoperation. These recorded trajectories can then be saved in the motion library and reused, promoting task-specific autonomy and bolstering the system's learning efficiency.
The contributions of this paper are as follows:
\begin{enumerate}
\item A GPT4-based LLM system has been developed to facilitate task planning for complex, long-horizon tasks. The LLM selects motion functions from the motion library according to the natural language commands. The selected functions are then integrated with environmental information, perceived through a YOLOv5-based perception module, enabling the autonomous execution of a wide range of tasks.
\item The proposed LLM system adopts a hierarchical planning framework by utilizing the prompt function of GPT4.
The LLM can dissect complex tasks into sub-tasks, further break down the sub-tasks into several motion functions, and execute each motion function sequentially.
\item A teleoperation-based HRC framework for motion demonstration is proposed. By integrating with DMP, the framework allows the LLM-based robot to learn from human demonstrations, thereby augmenting its motion capabilities.
\item The proposed HRC framework significantly enhances the capabilities of the LLM-based system in executing complex tasks. 
These tasks, often requiring intricate trajectory planning and reasoning over environments, can be efficiently accomplished through the incorporation of human demonstrations.
\end{enumerate}
 

 

\section{Related Works}
\subsection{Robotics Manipulation with Natural Language}
LLMs show great potential in the study of autonomous robotics control. 
Liang et al. \cite{liang2023code} proposed ``Code as Policies", a method using the language model to generate robot policy code from natural language commands and controlling the robot motion by executing the generated code.
Valuable enhancements and optimizations have been carried out based on the contributions of Liang and other related pioneers. In \cite{10341846}, the authors present the SoftGPT, a language model that combines graph representations and LLM-based dynamics for efficient soft object manipulation and learning from human demonstrations in domestic scenes. Besides, a multisensory approach is proposed in \cite{10160906} to enhance the robot's understanding and execution of natural language instructions for robot manipulation. We also utilize the LLM to assist in generating the necessary code for robot operation. In this case, the LLM is solely responsible for selecting the motion functions.

\subsection{VR-based Teleoperation System}
In 1940, the first robot teleoperation system was proposed by Goertz \cite{goertz1964manipulator}. Nowadays, tons of robot teleoperation systems with great manipulability and intelligence have been developed. Nakanishi et al. \cite{nakanishi2020towards} proposed an intuitive teleoperation system using a VR device. In addition, Zhu et al. present a force feedback system and a shared control framework to improve the transparency and manipulability of teleoperation \cite{zhu2020enhancing,10173494}. In \cite{lenz2023nimbro}, the authors introduce a robotic avatar system featuring immersive 3D visualization and advanced force-feedback telemanipulation.
In this paper, we use the teleoperation system as an instructive bridge between the human supervisor and the LLM-based robot manipulation system. 

\subsection{Dynamic Movement Primitives for Trajectory Learning}
Dynamic Movement Primitives (DMP) is a mathematical framework used in robotics and machine learning to model and reproduce complex joint behaviors. This approach was first proposed in \cite{schaal2006dynamic}, and updated by \cite{ijspeert2013dynamical}, the authors introduce DMP as a method for modeling complex behaviors in robotics, demonstrating their flexibility, adaptability, and potential for integration with statistical learning methods. 

After plenty of years of development, DMP has been considered a classic approach in the field of robotic trajectory learning. In 2013s, Paraschos et al. \cite{paraschos2013probabilistic} proposed an improved movement primitives approach named Probabilistic movement primitives (ProMP), which allow for blending motions, adapting to task variables, and co-activating movement primitives with advantages of probabilistic functions. Recently, in \cite{10050558}, the authors propose ProDMP, a method combining DMP and ProMP for smooth, statistics-informed trajectory generation, integrated with a deep neural network for enhanced learning and functionality.
In our architecture, we combine DMP with our teleoperation system to accomplish the one-shot task instructions.
\section{Methods}
Our proposed method combines an LLM with environmental information within the Robot Operating System (ROS) to construct an LLM-based autonomous system. To enhance the capabilities of the LLM-based system in executing complex tasks, we also adopt an HRC method to guide robot motion with human demonstration. This integration enables the translation of human commands into specific robotic motions. The system utilizes two primary libraries for motion execution: the basic library and the DMP library.
\begin{itemize}
    \item[] \textbf{Basic library}: This library includes pre-programmed motion functions. The LLM selects motion functions based on the task requirements and integrates these functions with environmental information to generate Pythonic code. This code controls the robot to perform motions corresponding to the tasks.
    \item[] \textbf{DMP library}: This library, developed using DMP technology, stores updated motion function sequences for sub-tasks (short-horizon tasks). These motion function sequences are updated through the user interface which leverages teleoperation.
\end{itemize}

To accomplish long-horizon tasks, the LLM can extract updated one-shot sub-tasks from the DMP library and combine them with other zero-shot sub-tasks composed by the motion functions from the basic library.

\subsection{Large Language Model for Autonomous Robot Manipulation}
In our methodology, we employ the GPT-4 Turbo network (the temperature is set to 0) to formulate the robot control hub. This hub is dedicated to converting user input into motion functions that can be integrated into Pythonic code. To ensure the model's output meets specific requirements, a special customization process is undertaken by providing the prompt. To clarify the LLM's duty, we provide the prompt ``Your mission is to control the robot to complete the task assigned by the user, utilizing the motion functions from the basic library". Also, the prompts introducing the components of the basic library are provided as follows:
``(i) \textquotesingle \textit{move\_to\_position()}\textquotesingle\ controls the robot end effector to reach the desired pose by updating joint angles through the calculation of inverse kinematics.
(ii) \textquotesingle \textit{base\_cycle\_move()}\textquotesingle\ controls the base of the robot to make a circle movement according to the given angle and radius.
Additionally, the function includes the motion to fully open the door.
(iii) \textquotesingle \textit{close\_move()}\textquotesingle\ is used to close doors with vertical door axes. The scope of motion is decided by the size of the object.
(iv) \textquotesingle \textit{gripper\_control()}\textquotesingle\ controls the opening and closing of the gripper and can be used to control the degree of opening and closing depending on the size of the object."
Through this typical prompt, the LLM can generalize and perform similar tasks without relying on additional related prompts, illustrating that while prompt design is crucial, it does not constitute the predominant workload in operations.

Additionally, to ensure the LLM executes motion functions properly, providing detailed prompts is important. For example, customization inputs should include instructive examples, such as ``When I ask you to put the apple on the plate, you can arrange motion functions like this \textquotesingle move\_to\_position(init)\textquotesingle, \textquotesingle move\_to\_position(apple)\textquotesingle, \textquotesingle gripper\_control(close\_low)\textquotesingle, \textquotesingle move\_to\_position(init)\textquotesingle, \textquotesingle\ move\_to\_-\\position(plate)\textquotesingle, \textquotesingle\ gripper\_control(open)\textquotesingle, \textquotesingle\ move\_to\_position(-\\init)\textquotesingle.''

Moreover, to address ambiguity or vagueness in commands, the system can be enhanced with additional prompts for customization. To give an example of such prompts, ``If multiple objects share the same name as the target, you could cue the user by asking, `There are multiple objects share the same name, which one do you prefer?'"

Based on these customizations, the LLM is enabled to select and utilize motion functions from the basic library in response to user requirements, achieving 99.4\% executability (Table \ref{tab:success_rates}).

Besides, a hierarchical strategy of task planning is applied. The hierarchical task planning strategy in our system enables handling complex, long-horizon tasks in the real world. For example, for a task like heating food, the LLM devises a detailed plan broken down into specific sub-tasks with motion functions, but not a long sequence of motion functions is prone to errors.
To classify different tasks, the prompt “As you receive a task, first consider if the task can be divided into sub-tasks, if so divide them into sub-tasks with motion functions, if not, directly output the motion functions” is given to the LLM. As illustrated in Fig. 2, the LLM translates short-horizon tasks directly into motion functions \(MoF(t)\), whereas long-horizon tasks are divided into sub-tasks, each comprising multiple motion functions \(\sum^N_iMoF(t_i)\).
\begin{figure}[t]
\vspace{2mm}
    \centering
    \includegraphics[width=0.9\linewidth]{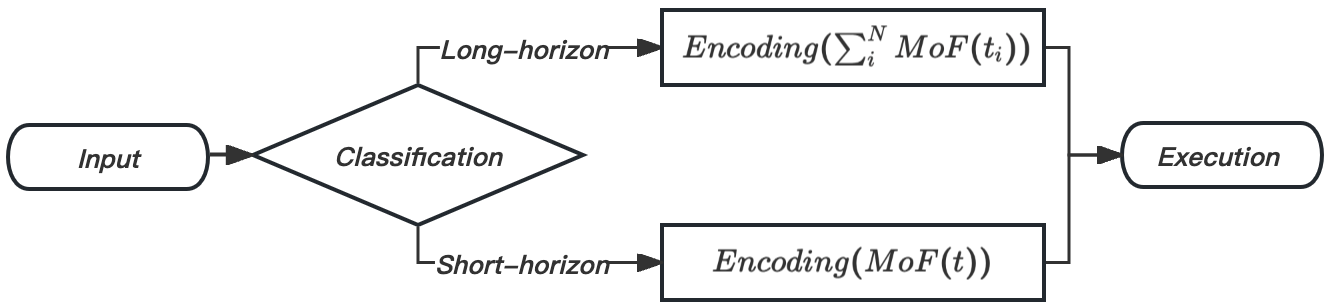}
    \captionsetup{skip=2pt}
    \caption{Illustration of how the LLM understands, classifies, decomposes, and executes different tasks.}
    \label{fig:long-horizon}
 \vspace{-5mm}
\end{figure}
\begin{figure*}[t]
\vspace{1mm}
    \centering
    \includegraphics[width=0.8\textwidth]{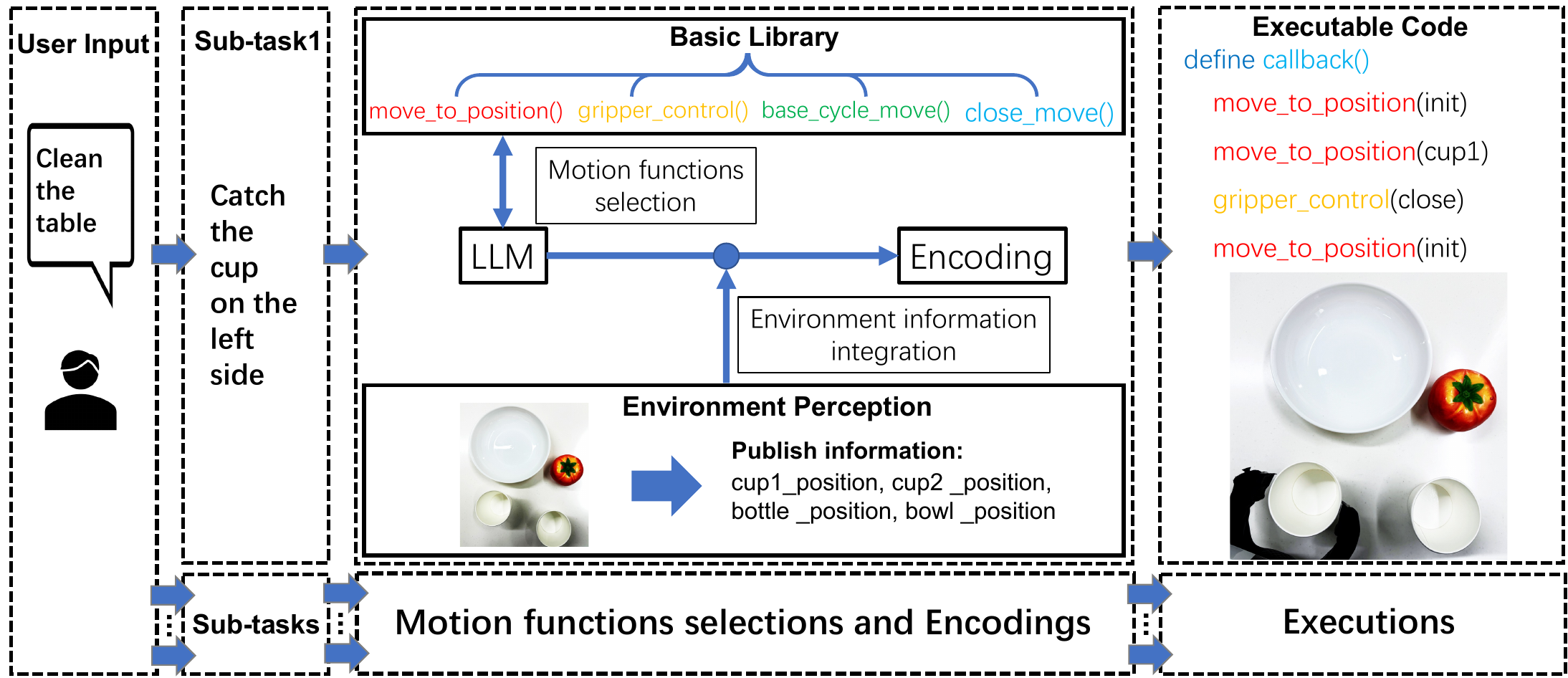}
    \captionsetup{skip=0pt}
    \caption{An overview of LLM-based autonomy for a real-world long-horizon task. This process encompasses sub-task identification, motion function selection from the basic library, environment perception integration, and executable code generation.}
    \label{fig:autonomy}
 \vspace{-5mm}
\end{figure*}

In practice, user commands for long-horizon tasks, are illustrated in Fig.3 and processed through the following steps:
\begin{enumerate}
    \item The user issues a command (e.g., Clean the table).
    \item The LLM interprets this command and classifies it to a long-horizon task. Based on the classification, this task can be divided into several sub-tasks, such as the task Put the left cup in storage.
    \item For the first sub-task, the LLM parses and maps it to the motion functions contained within the basic library. 
    \item The environmental perception system analyzes sensory data and publishes the spatial coordinates of relevant entities within the operational environment, such as bottle1\_position and bowl\_position in Fig. 3.
    \item The processed sensory data and the motion function sequence are integrated into an executable Pythonic code. By executing this Pythonic code, the robot is controlled to fulfill the first sub-task. The remaining sub-tasks will be executed following the aforementioned process.
\end{enumerate}

\subsection{Environmental Perception}
In the environmental perception part, we extract the target objects' names, quantities, and coordinates in the image by the robot's Xtion depth camera based on YOLOv5. The names and quantities of these objects are used to dynamically update the prompts for the LLM, which is used for LLM's motion planning in response to user commands. The coordinates of the target to the depth camera can be calculated as follows: 
\begin{equation}
\boldsymbol{P}^{\textit{c}}_{\textit{obj}} = d \cdot \boldsymbol{K}^{-1} \cdot \boldsymbol{i}
\end{equation}
where \(\boldsymbol{i} = \begin{bmatrix} u & v & 1 \end{bmatrix}^T\) represents the homogeneous coordinates in the image, corresponding to the target object's image coordinate point \((u, v)\). The term \(d\) denotes the depth value at the point \((u, v)\) collected from the depth map, and \(\boldsymbol{K}\) refers to the intrinsic matrix of the depth camera. The position of the object relative to the depth camera is denoted by \(\boldsymbol{P}^{\textit{c}}_{\textit{obj}}\).

The world coordinate of the target is given by the following transformation:
\begin{equation}
\boldsymbol{P}^{\textit{w}}_{\textit{obj}} = \boldsymbol{T}_{\textit{c}}^{\textit{w}} \cdot \boldsymbol{P}^{\textit{c}}_{\textit{obj}}
\end{equation}
where \(\boldsymbol{P}^{\textit{w}}_{\textit{obj}}\) represents the position of the object in world coordinates, and \(\boldsymbol{P}^{\textit{c}}_{\textit{obj}}\) denotes its position in camera coordinates. The transformation matrix \(\boldsymbol{T}_{\textit{c}}^{\textit{w}}\) encapsulates the rotation \(R_{\textit{c}}^{\textit{w}}\) and translation \(t_{\textit{c}}^{\textit{w}}\) matrix from the camera coordinate system to the world coordinate system. 

The calculated target positions are continuously published and labeled with their names. For the ongoing task, the LLM integrates these calculated positions with motion functions, guiding the robot's movements as illustrated in Fig. 3.

On the other hand, to efficiently handle complex scenarios involving multiple same-name targets and pre-tasks (obstacle removal), two specialized algorithms are proposed. 
(i) Algorithm 1 streamlines the object selection in cluttered environments. It processes the same-named objects detected by YOLOv5, sorting and labeling them from spatial left to right within each object category. By integrating the detected objects and their quantities into the prompt, the LLM can interpret commands for specific objects. For example, the task ``Pick the middle cup" is achieved by picking the appropriately labeled object `cup2', with the position information provided by Algorithm 1 (e.g., `cup2' in a sorted sequence of three cups).
(ii) Algorithm 2 addresses obstacle removal before task execution.
When the LLM executes a task, it extracts the targets' names involved and forwards their names to Algorithm 2. Algorithm 2 then evaluates the workspace of each target, identifying any obstacles and returning a list of the obstacles to the LLM. Upon receiving this list, the original task is temporarily paused and the LLM initiates a new task to remove the identified obstacles. Only after these obstacles are successfully cleared does the LLM resume the execution of the planned motion sequence for the primary objective.
\begin{algorithm}
\caption{Sort and Label Objects Detected}
\small
\label{alg:1}
\begin{algorithmic}[1]
\State Input: List of objects $O$ detected by YOLOv5
\State Output: List of labeled objects $L$
\State Initialize $L = []$
\For{each unique object name $n$ in $O$}
    \State Filter objects in $O$ with name $n$, store in $O_n$
    \State Sort $O_n$ from left to right based on their positions, store in $O_{n\_sorted}$
    \For{each object $o$ in $O_{n\_sorted}$}
        \State Label $o$ with its index in $O_{n\_sorted}$
        \State Add labeled $o$ to $L$
    \EndFor
\EndFor
\State Return $L$
\end{algorithmic}
\end{algorithm}
\begin{algorithm}
\caption{Identify Obstacles for Task Execution}
\small
\label{alg:2}
\begin{algorithmic}[1]
\State Input: Target coordinates $TCoord$, robot base coordinates $RCoord$, workspace objects $WObjs$
\State Output: Coordinates sequence of obstacles $ObsCoord$
\State Initialize $ObsCoord$ as an empty list
\State Define an equilateral triangular detection space with one vertex at $RCoord$ and the base midpoint at $TCoord$

\If{$WObjs$ is empty}
    \State Return `No objects'
\EndIf
\For{each object $Obj$ in $WObjs$}
    \State Check if $Obj$ is within the triangular detection space
    \If{$Obj$ is inside}
        \State Add $Obj$'s coordinates to $ObsCoord$
    \EndIf
\EndFor
\State Return $ObsCoord$
\end{algorithmic}
\end{algorithm}
\vspace{-2mm}

 \begin{figure*}[t]
 \vspace{1.3mm}
    \centering
    \includegraphics[width=0.9\linewidth]{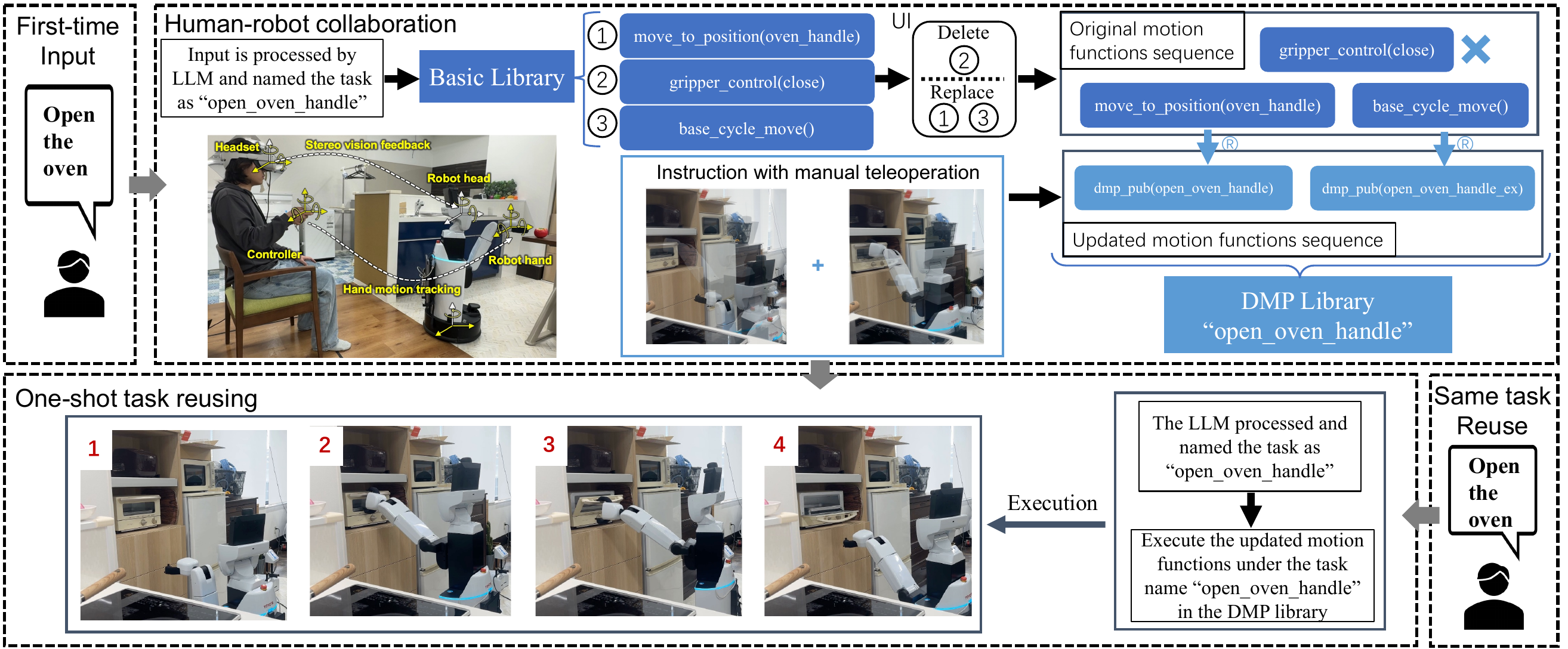}
    \vspace{-3mm}
    \caption{An overview of the LLM-based autonomy with Human-Robot Collaboration in sub-task (short-horizon task). The LLM processes user input to select motion functions from the basic Library. These selected motions are subsequently modified through the user interface with teleoperation. The updated motion functions are stored in the DMP Library with a specific name such as ``open\_oven\_handle" (The LLM captures the action ``open" and the target ``oven\_handle", then integrates them as ``open\_oven\_handle") for future application (same task re-input or reusing in the long-horizon task), resulting in successful one-shot task execution.}
    \label{fig:hrc}
  \vspace{-5mm}
\end{figure*}
\subsection{Human-Robot Collaboration}
Although LLMs can assist in basic tasks when integrated with environmental perception, their limitations become apparent in complex, real-world scenarios. For example, while the LLM successfully operates a microwave with a normal design, it could struggle with opening an oven door featuring a horizontal axis design. This difficulty stems from the LLM's limited capabilities in code generation, particularly in handling a constrained motion library and understanding complex trajectories. Furthermore, even though GPT-4 can interpret environments through image input, it still relies on human guidance for accurate decision-making in diverse and unfamiliar situations.

To address these challenges, we advocate for a synergistic framework of human-robot collaboration that is grounded in LLM-based autonomy. In this proposed framework, users are capable of proactively interceding and seamlessly instructing the robot by leveraging teleoperation. The instructed motions can be used to replace the original motion functions in the basic library and the updated motion functions sequences are saved in the DMP library through the user interface shown in Fig. 5.

The teleoperation system \cite{nakanishi2020towards} includes a headset and a controller for the operator. The headset provides stereo-vision feedback, replicating a real-world view as seen through the robot's eyes, and mirrors the user's head movements in the robot's head movements. The controller, held in the user’s hand, maps the user's hand movements to the robot's hand movements. These setups ensure that operators can intuitively guide the robot, as illustrated in Fig. 4.

These custom motions, instructed by the operator through the teleoperation system, are then captured and integrated into the robot's capabilities using DMP technology, as updated by Ijspeert \cite{ijspeert2013dynamical}. DMP technology is crucial for capturing and storing these tailored motion trajectories in the DMP library, thereby enriching the robot's repertoire for diverse tasks.
The core principle of DMP is encapsulated in the following equations:
\begin{equation}
    \ddot{y} = \alpha (\beta (g - y) - \dot{y}) + f(x) (g - y)
\end{equation}
where \(y\) represents the system's state, \(g\) is the goal, \(\alpha\) and \(\beta\) are the constant gain terms. This equation describes the behavior of the trajectory. For the forcing term \(f(x)\):
\begin{equation}
    f(x) = \frac{\sum_{i=1}^{N} \psi_i w_i x}{\sum_{i=1}^{N} \psi_i}
\end{equation}
where \(\psi_i\) are Gaussian basis functions, \(w_i\) the corresponding weights, and \(x\) is the phase variable from the canonical system that decays non-linearly. The weights \(w_i\) are derived from the training trajectory using weighted linear regression, and \(N\) represents the number of Gaussian basis functions, fixed at 15 in our experiments.
DMP achieves trajectory storage and replay by learning the weights \( w_i \) from a demonstrated trajectory achieved by teleoperation. Once learned, these weights define the shape of the forcing function \( f(x) \), which, when combined with the spring-damper system in Equation (3), produces the desired movement trajectory. By varying the goal \( g \), start position \( y \), and phase \( x \), the DMP can generate different trajectories while still maintaining the learned movement pattern.
\begin{figure}[t]
    \centering
    \includegraphics[width=0.79\linewidth]{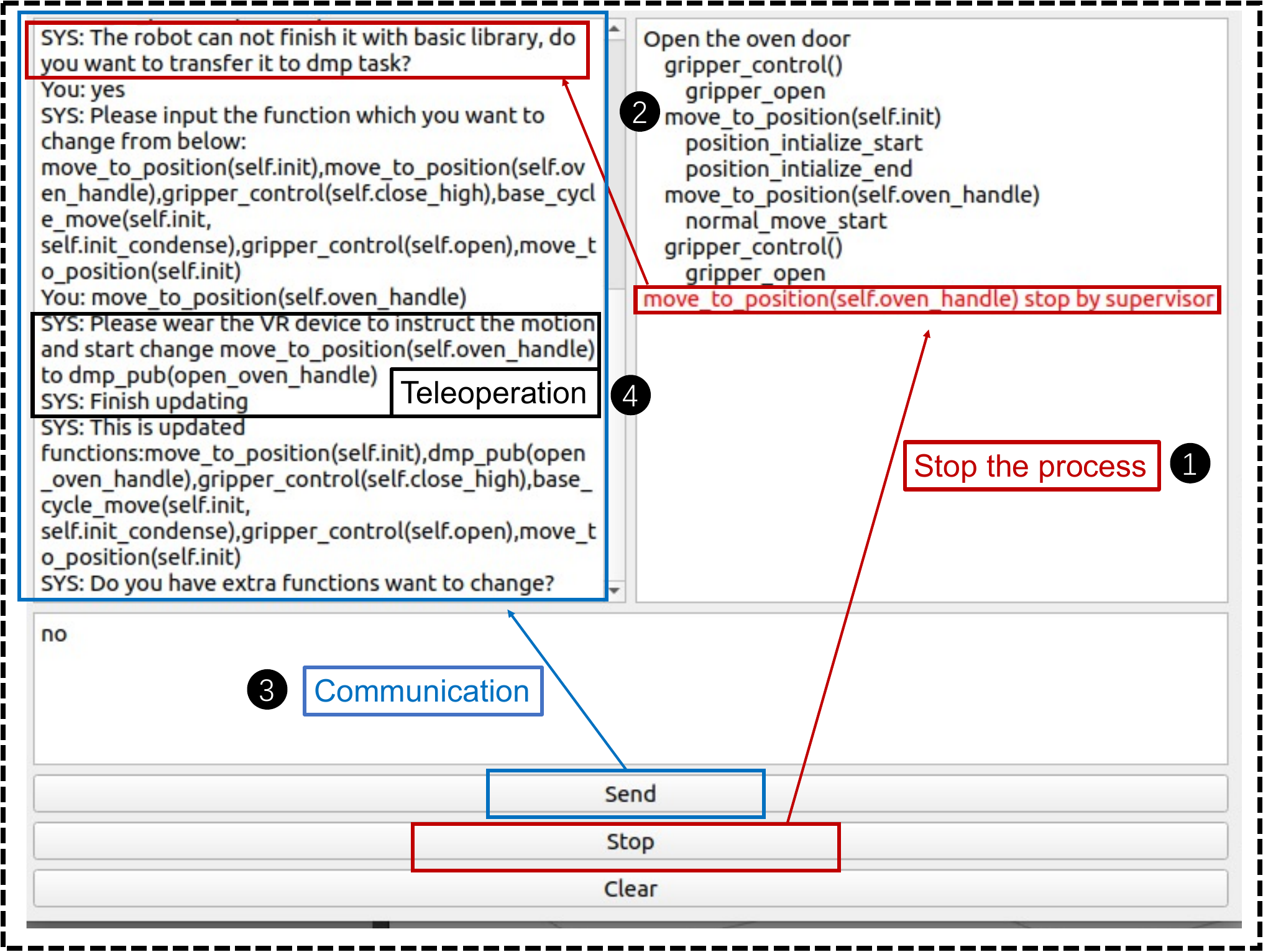}
    \captionsetup{skip=1pt}
    \caption{An overview of the usage of the user interface. A user interface is used for supervising and intervening in a robot’s task sequence, including the communication between the user and the system and the steps for modifying robotic motions through teleoperation.}
    \label{fig:ui}
 \vspace{-6mm}
\end{figure}

By leveraging teleoperation for trajectory recording, the system enables an LLM-based robot to improve previously problematic tasks by accessing human-defined motion trajectories stored in the DMP library.

The HRC process in the real world, which is conducted through the user interface shown in Fig. 5, is illustrated in Fig. 4, following the steps as follows:
\begin{enumerate}
    \item The user begins the process by issuing a command, such as ``Open the oven".
    \item The LLM could process the command, map it into a set of basic motions that exist in the robot's basic library, and integrate it with environment information which is the same process in Fig. 3.
    \item However, in Fig. 4, the original motion sequence, suitable for a door with a vertical axis, will fail to open the horizontally hinged oven door. The user notices it, stops the process, and updates motion functions via the user interface and teleoperation shown in Fig. 5.
    \item The updated new motion functions sequence is then saved in the DMP library and the new motion trajectory instructed by manual teleoperation can be called by the function \textquotesingle dmp\_pub\textquotesingle. These data are preserved in the DMP library named ``open\_oven\_handle" which is named by the LLM.
    \item When the task ``Open the oven" is retrieved, the LLM names the task as ``open\_oven\_handle" again, extracts the updated motion functions sequence from the task’s segment in the DMP library, and enables the robot to perform the task as depicted in Fig. 4.
\end{enumerate}

\section{Experiments and Discussion}
The HRC framework based on the LLM has undergone real-world testing on the Human Support Robot (HSR) from Toyota. 
The HSR has a total of 10-Degree-of-Freedom (DoF), including a 3-DoF mobile base, a 5-DoF arm (4-DoF for rotating joints and 1-DoF for a torso lift joint) with a gripper, and a 2-DoF head. 
Oculus’s VR device is used for teleoperation. The experiment objects are shown in Fig. 6.

\begin{figure}[t]
\vspace{2mm}
\centering

\subfloat[HSR joint configuration]{%
   \includegraphics[width=0.19\textwidth]{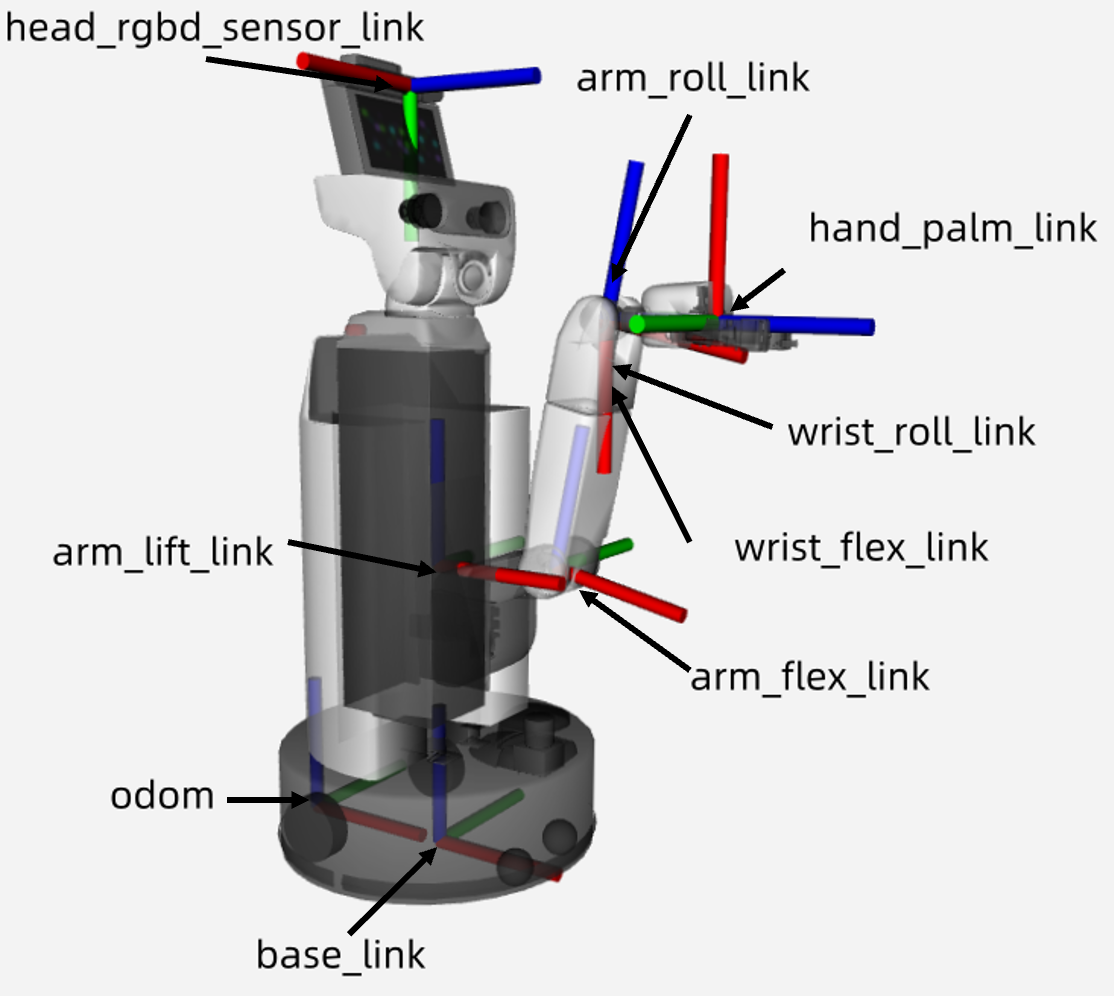}
   \label{fig:joint}
}
\hspace{3pt}
\subfloat[Environment setup]{%
   \includegraphics[width=0.225\textwidth]{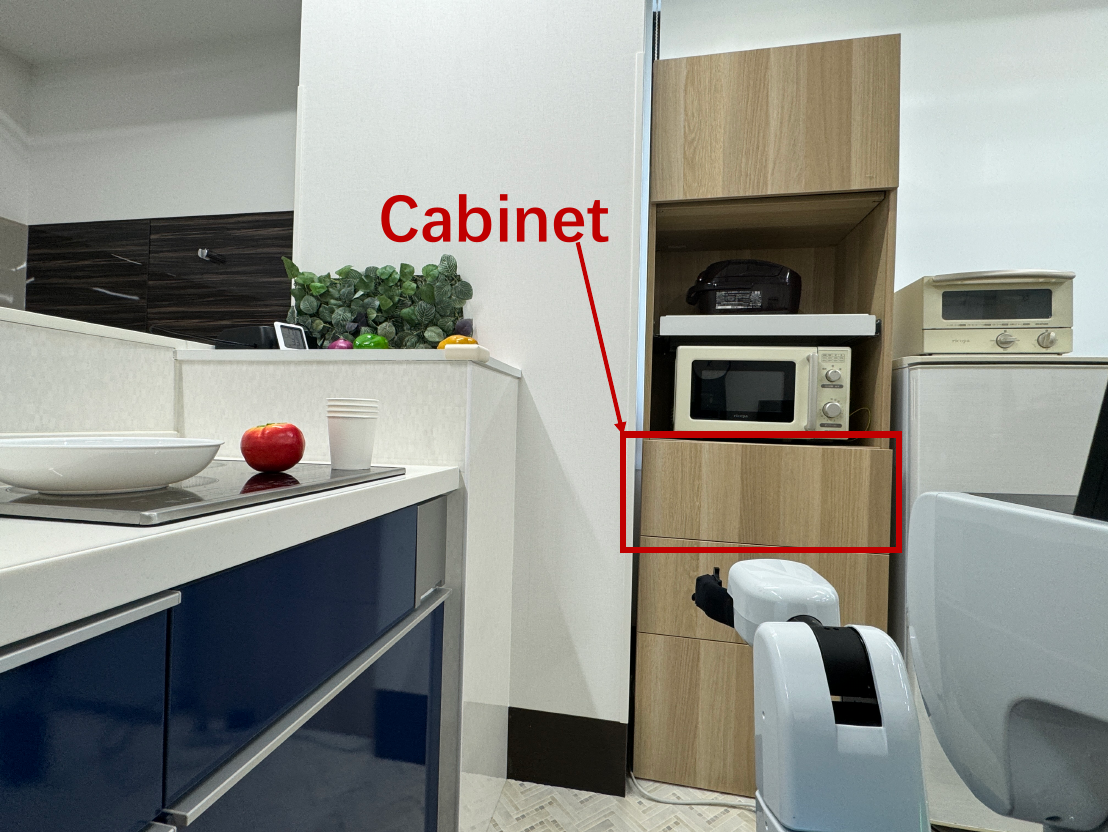}
   \label{fig:environment}
}
\vspace{3pt}
\subfloat[Oven]{%
   \includegraphics[width=0.208\textwidth]{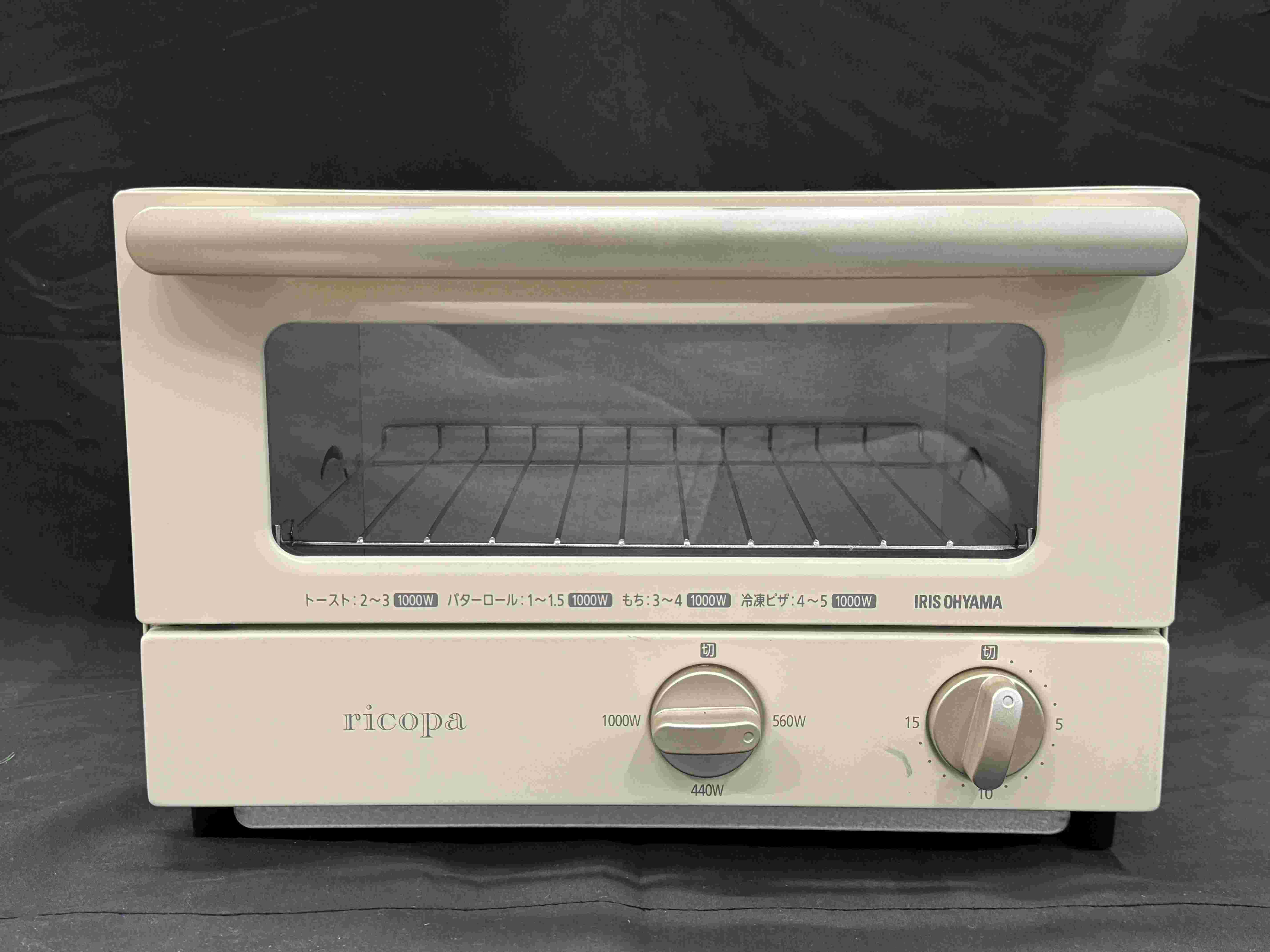}
   \label{fig:oven}
}
\hspace{3pt}
\subfloat[Microwave]{%
   \includegraphics[width=0.208\textwidth]{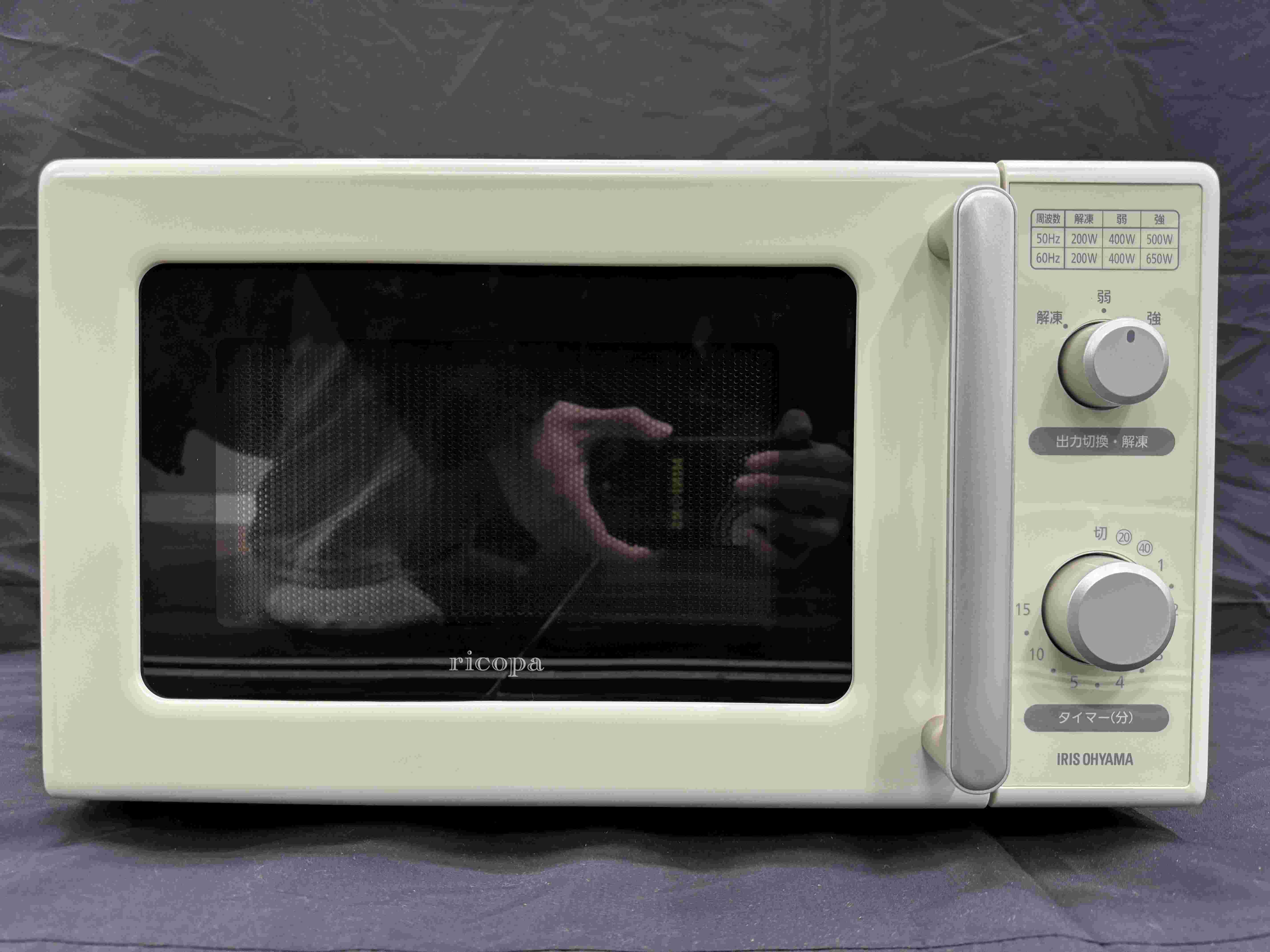}
   \label{fig:microwave}
}
\vspace{3pt}
\subfloat[Apple]{%
   \includegraphics[width=0.132\textwidth]{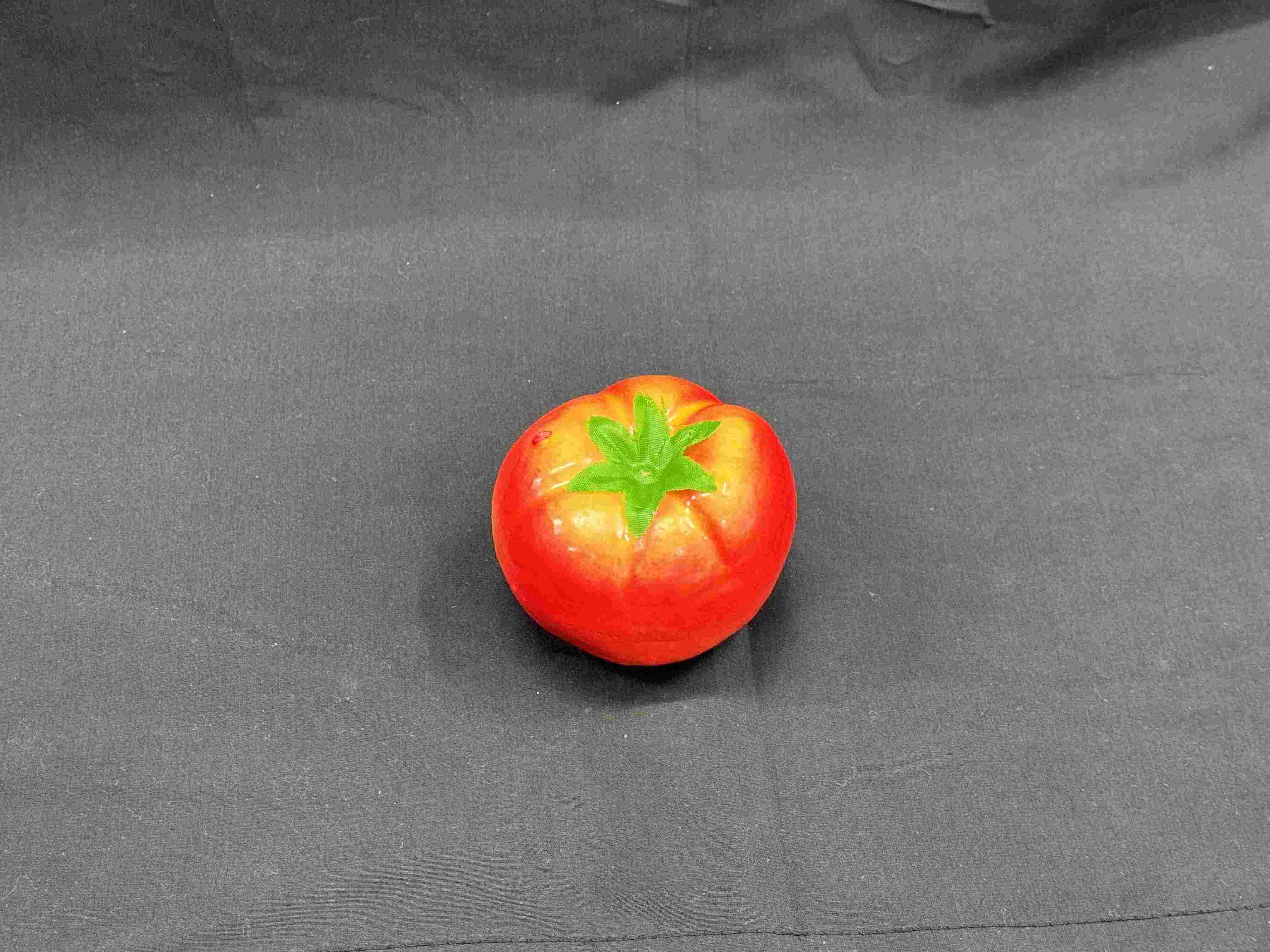}
   \label{fig:apple}
}
\hspace{3pt}
\subfloat[Bowl]{%
   \includegraphics[width=0.132\textwidth]{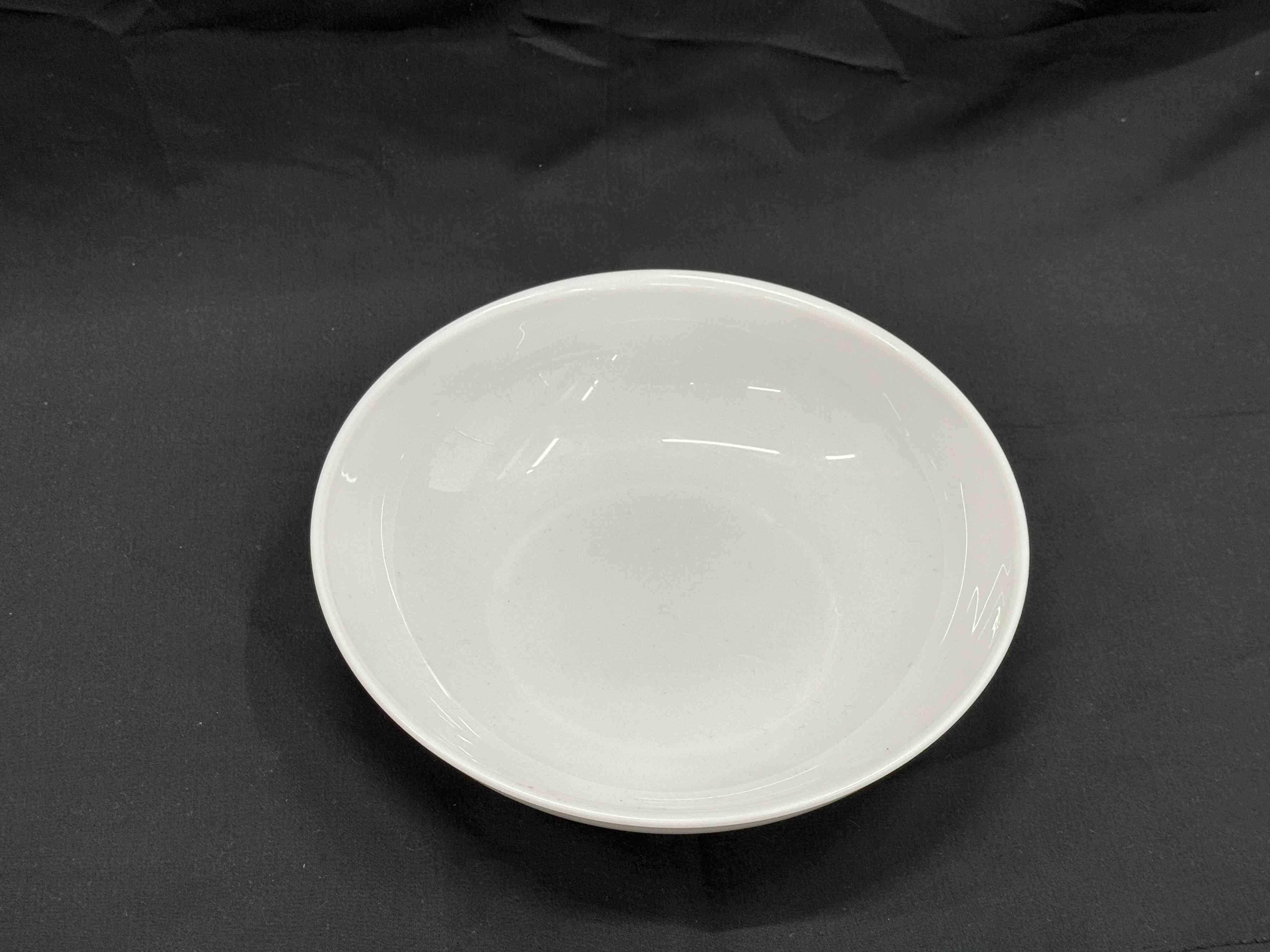}
   \label{fig:bowl}
}
\hspace{3pt}
\subfloat[Cups]{%
   \includegraphics[width=0.132\textwidth]{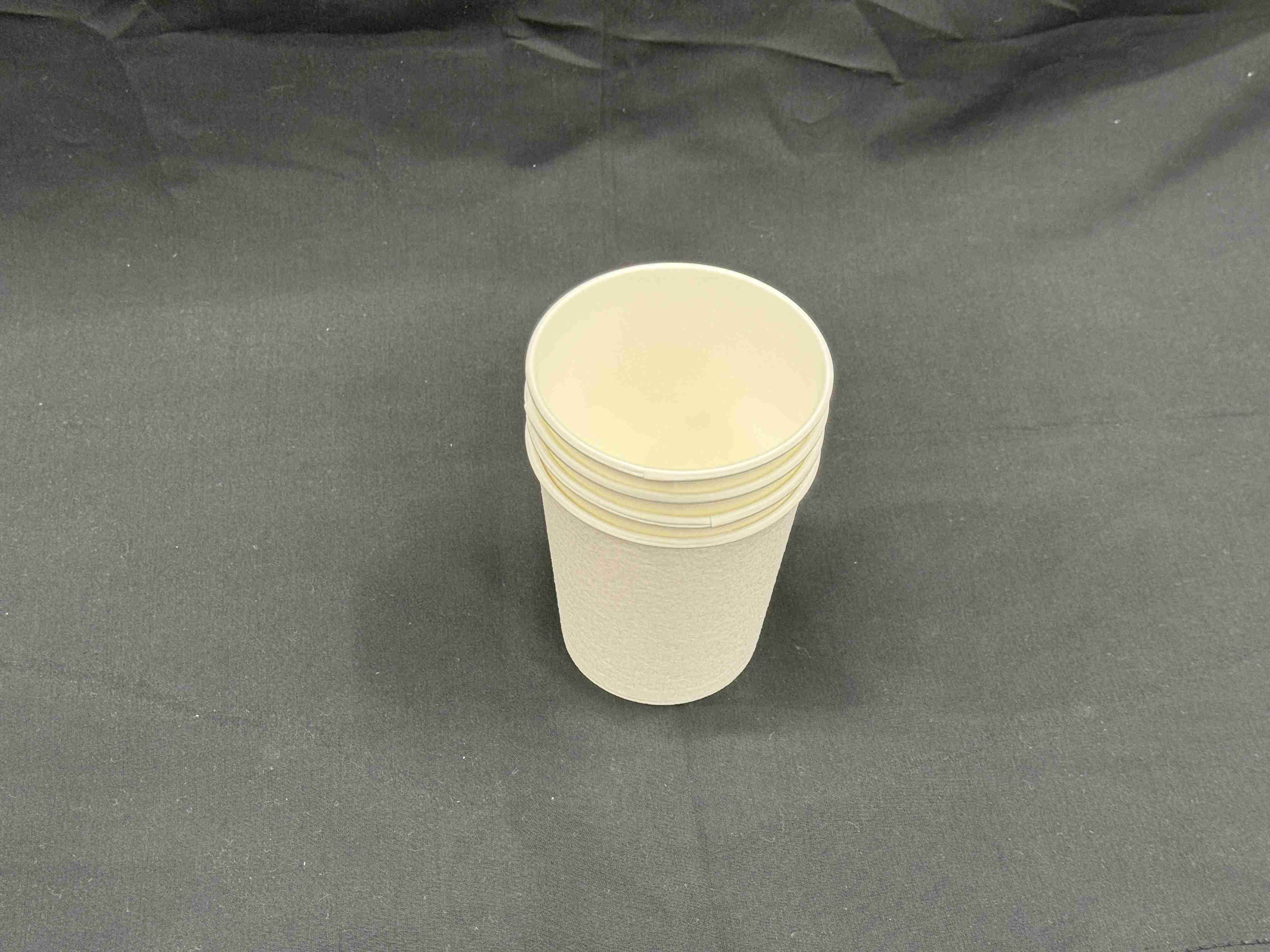}
   \label{fig:cups}
}

\caption{HSR joint configuration, experiment environment setup, and objects utilized in the experiments}
\label{fig:full_config}
 \vspace{-4mm}
\end{figure}

The experiments were set in a well-lit kitchen environment as shown in Fig. 6 (b), where the robot was tasked with performing a variety of domestic tasks initiated through natural language commands provided by users. 
The experiments aim to evaluate the performance of the system for routine zero-shot tasks and more intricate and specialized one-shot tasks.
\subsection{Zero-shot Basic Tasks}
In this part, we conducted two basic tasks {Put\&Stack, and Open}, and two long-horizon tasks {Warm up and Clean}.

\subsubsection{Put\&Stack}
The Put motions position an item relative to a target and are of two types: above or inside the target location. The customized LLM adds a suffix to specify the target context—\textquotesingle \_above\textquotesingle\ for placement 10mm higher than the target, and \textquotesingle \_inside\textquotesingle\ for insertion, such as \textquotesingle object\_above\textquotesingle.
The motion functions for Put\&Stack are as follows:
\begin{codebox}
\textcolor{codegreen}{\footnotesize \texttt{\# Move to the first object's position}}\\
\colorbox{codeblue}{%
    \begin{minipage}{\dimexpr\textwidth-6pt\relax}
        \footnotesize \texttt{move\_to\_position(object1)}
    \end{minipage}%
}\\
\textcolor{codegreen}{\footnotesize \texttt{\# Close gripper}}\\
\colorbox{codeblue}{%
    \begin{minipage}{\dimexpr\textwidth-6pt\relax}
        \footnotesize \texttt{gripper\_control(close)}
    \end{minipage}%
}
\textcolor{codegreen}{\footnotesize \texttt{\# Move to the second object's above position}}\\
\colorbox{codeblue}{%
    \begin{minipage}{\dimexpr\textwidth-6pt\relax}
        \footnotesize \texttt{move\_to\_position(object2\_above/inside)}
    \end{minipage}%
}\\
\textcolor{codegreen}{\footnotesize \texttt{\# Open gripper}}\\
\colorbox{codeblue}{%
    \begin{minipage}{\dimexpr\textwidth-6pt\relax}
        \footnotesize \texttt{gripper\_control(open)}
    \end{minipage}%
}
\end{codebox}
\vspace{-1.5mm}

\subsubsection{Open}
The effect of this motion Open is opening the door, but only for the door with the axis perpendicular to the ground.
\begin{codebox}
\textcolor{codegreen}{\footnotesize \texttt{\# Move to the door handle's position}}\\
\colorbox{codeblue}{%
    \begin{minipage}{\dimexpr\textwidth-6pt\relax}
        \footnotesize \texttt{move\_to\_position(microwave\_handle)}
    \end{minipage}%
}\\
\textcolor{codegreen}{\footnotesize \texttt{\# Close gripper}}\\
\colorbox{codeblue}{%
    \begin{minipage}{\dimexpr\textwidth-6pt\relax}
        \footnotesize \texttt{gripper\_control(close)}
    \end{minipage}%
}
\textcolor{codegreen}{\footnotesize \texttt{\# Open the door}}\\
\colorbox{codeblue}{%
    \begin{minipage}{\dimexpr\textwidth-6pt\relax}
        \footnotesize \texttt{base\_cycle\_move(radius\_door2axis)}
    \end{minipage}%
}\\
\textcolor{codegreen}{\footnotesize \texttt{\# Open gripper}}\\
\colorbox{codeblue}{%
    \begin{minipage}{\dimexpr\textwidth-6pt\relax}
        \footnotesize \texttt{gripper\_control(open)}
    \end{minipage}%
}
\end{codebox}
\vspace{-1.5mm}
\subsubsection{Close}
This motion is used to close the door, also only for the door with the axis perpendicular to the ground.
\begin{codebox}
\textcolor{codegreen}{\footnotesize \texttt{\# Close the door}}\\
\colorbox{codeblue}{%
    \begin{minipage}{\dimexpr\textwidth-6pt\relax}
        \footnotesize \texttt{close\_move(object)}
    \end{minipage}%
}
\end{codebox}
\vspace{-1.5mm}
\subsubsection{Power on}
This motion includes the function to move to the target and rotate the waist to power the device.
\begin{codebox}
\textcolor{codegreen}{\footnotesize \texttt{\# Move to the object}}\\
\colorbox{codeblue}{%
    \begin{minipage}{\dimexpr\textwidth-6pt\relax}
        \footnotesize \texttt{move\_to\_position(microwave\_knob)}
    \end{minipage}%
}
\textcolor{codegreen}{\footnotesize \texttt{\# Rotate waist to power on the object}}\\
\colorbox{codeblue}{%
    \begin{minipage}{\dimexpr\textwidth-6pt\relax}
        \footnotesize \texttt{rotate\_waist(degree)}
    \end{minipage}%
}
\end{codebox}
\vspace{-1.5mm}
\subsubsection{Warm up}
This task is a long-horizon task to warm up the apple which contains several sub-tasks.
\begin{codebox}
\textcolor{codegreen}{\footnotesize \texttt{\# Generate sub-tasks}}\vspace*{1mm}\\
\colorbox{codeblue}{%
    \begin{minipage}{\dimexpr\textwidth-6pt\relax}
       \footnotesize  \texttt{sub-tasks = [\textquotesingle open the microwave\textquotesingle , \textquotesingle put the apple into the microwave\textquotesingle , \textquotesingle close the microwave\textquotesingle , \textquotesingle power on the microwave\textquotesingle ]}
    \end{minipage}%
}
\textcolor{codegreen}{\footnotesize \texttt{\# Execute the functions of the sub-task in order}\vspace{1mm}}\\
\colorbox{longhri}{%
    \begin{minipage}{\dimexpr\textwidth-6pt\relax}
        \footnotesize \texttt{for sub-task in sub-tasks:\\
            \hspace*{0.5cm}execute(sub-task)}
    \end{minipage}%
}
\end{codebox}
\vspace{-1.5mm}
\subsubsection{Clean the table}
This task is a long-horizon task. The environmental information including the objects on the table is generated to integrate into executable code. In the below case, we assume that there are one cup and two bottles. 
\begin{codebox}
\textcolor{codegreen}{\footnotesize \texttt{\# Generate sub-tasks}}\vspace*{1mm}\\
\colorbox{codeblue}{%
    \begin{minipage}{\dimexpr\textwidth-6pt\relax}
        \footnotesize \texttt{sub-tasks = [\textquotesingle put the cup in the storage\textquotesingle , \textquotesingle put the first bottle in the storage\textquotesingle , \textquotesingle put the second bottle in the storage\textquotesingle} ]
    \end{minipage}%
}
\textcolor{codegreen}{\footnotesize \texttt{\# Execute the functions of the sub-task in order}\vspace{1mm}}\\
\colorbox{longhri}{%
    \begin{minipage}{\dimexpr\textwidth-6pt\relax}
        \footnotesize \texttt{for sub-task in sub-tasks:\\
            \hspace*{0.5cm}execute(sub-task)}
    \end{minipage}%
}
\end{codebox}
\vspace{-1.5mm}
\subsection{One-shot DMP-based Tasks}
In this section, we conducted two one-shot DMP-based tasks and a long-horizon task ``Roast apple (HRC)".
\subsubsection{Open the oven}
The motion function \textquotesingle base\_cycle\_mov-\\e()\textquotesingle\ in our basic library is not effective for the door with the horizontal axis. To address this, we teach it through teleoperation and save it in the DMP library. 
\begin{codebox}
\textcolor{codegreen}{\footnotesize \texttt{\# Move to the oven's handle with updated pose}}\\
\colorbox{codeblue}{%
    \begin{minipage}{\dimexpr\textwidth-6pt\relax}
       \footnotesize  \texttt{move\_to\_position(oven\_handle)}
    \end{minipage}%
}\\
\hspace*{2cm}$\downarrow$ \texttt{Replace}\\
\colorbox{dmpcolor}{%
    \begin{minipage}{\dimexpr\textwidth-6pt\relax}
        \footnotesize \texttt{dmp\_publish(open\_oven\_handle)}
    \end{minipage}%
}
\textcolor{codegreen}{\footnotesize \texttt{\# Delete the gripper\_control(close)}}\\
\colorbox{codeblue}{%
    \begin{minipage}{\dimexpr\textwidth-6pt\relax}
        \sout{\footnotesize \texttt{gripper\_control(close)}}
    \end{minipage}%
}\\
\textcolor{codegreen}{\footnotesize \texttt{\# Open the oven with updated trajectory}}\\
\colorbox{codeblue}{%
    \begin{minipage}{\dimexpr\textwidth-6pt\relax}
        \footnotesize \texttt{base\_cycle\_move(radius\_door2axis)}
    \end{minipage}%
}\\
\hspace*{2cm}$\downarrow$ \texttt{Replace}\\
\colorbox{dmpcolor}{%
    \begin{minipage}{\dimexpr\textwidth-6pt\relax}
        \footnotesize \texttt{dmp\_publish(open\_oven\_handle\_ex)}
    \end{minipage}%
}
\end{codebox}
\vspace{-1.5mm}
\subsubsection{Close the oven}
Due to the special construction of the oven in contrast to the microwave oven, closing the oven door also requires instruction to replace the motion function \textquotesingle close\_move(object)\textquotesingle.
\begin{codebox}
\textcolor{codegreen}{\footnotesize \texttt{\# Close the oven with new trajectory}}\\
\colorbox{codeblue}{%
    \begin{minipage}{\dimexpr\textwidth-6pt\relax}
        \footnotesize \texttt{close\_move(object)}
    \end{minipage}%
}\\
\hspace*{2cm}$\downarrow$ \texttt{Replace}\\
\colorbox{dmpcolor}{%
    \begin{minipage}{\dimexpr\textwidth-6pt\relax}
        \footnotesize \texttt{dmp\_publish(close\_oven)}
    \end{minipage}%
}
\end{codebox}
\vspace{-1.5mm}
\subsubsection{Open the cabinet}
The cabinet has a press-pull structure (To open it, the cabinet door must first be pushed inward.), which makes it impossible for the LLM to control the robot to open it in the zero-shot. (The position of the cabinet is determined by its relative location to the microwave.)
\begin{codebox}
\textcolor{codegreen}{\footnotesize \texttt{\#The new motion to press the cabinet}}\\
\colorbox{codeblue}{%
    \begin{minipage}{\dimexpr\textwidth-6pt\relax}
       \footnotesize  \texttt{move\_to\_position(cabinet)}
    \end{minipage}%
}\\
\hspace*{2cm}$\downarrow$ \texttt{Replace}\\
\colorbox{dmpcolor}{%
    \begin{minipage}{\dimexpr\textwidth-6pt\relax}
        \footnotesize \texttt{dmp\_publish(open\_cabinet)}
    \end{minipage}%
}
\textcolor{codegreen}{\footnotesize \texttt{\# Delete the gripper\_control(close)}}\\
\colorbox{codeblue}{%
    \begin{minipage}{\dimexpr\textwidth-6pt\relax}
        \sout{\footnotesize \texttt{gripper\_control(close)}}
    \end{minipage}%
}\\
\textcolor{codegreen}{\footnotesize \texttt{\# The new motion to pull the cabinet}}\\
\colorbox{codeblue}{%
    \begin{minipage}{\dimexpr\textwidth-6pt\relax}
        \footnotesize \texttt{base\_cycle\_move(radius\_door2axis)}
    \end{minipage}%
}\\
\hspace*{2cm}$\downarrow$ \texttt{Replace}\\
\colorbox{dmpcolor}{%
    \begin{minipage}{\dimexpr\textwidth-6pt\relax}
        \footnotesize \texttt{dmp\_publish(open\_cabinet\_ex)}
    \end{minipage}%
}
\end{codebox}
\vspace{-1.5mm}
\subsubsection{Roast the apple}
This task aims to determine whether the sub-tasks instructed by the DMP can be repurposed for relatively long-horizon tasks. 
\begin{codebox}
\textcolor{codegreen}{\footnotesize \texttt{\# Generate sub-tasks (one-shot DMP-based tasks are bold letter)}}\vspace*{1mm}\\
\colorbox{codeblue}{%
    \begin{minipage}{\dimexpr\textwidth-6pt\relax}
        \footnotesize \texttt{sub-tasks = [\textbf{\textquotesingle open the oven\textquotesingle }, \textquotesingle put the apple into oven\textquotesingle , \textbf{\textquotesingle close the oven\textquotesingle }, \textquotesingle power on the oven\textquotesingle ]}
    \end{minipage}%
}
\textcolor{codegreen}{\footnotesize \texttt{\# Execute the functions of the sub-task in order}\vspace{1mm}}\\
\colorbox{longhri}{%
    \begin{minipage}{\dimexpr\textwidth-6pt\relax}
        \footnotesize \texttt{for sub-task in sub-tasks:\\
            \hspace*{0.5cm}execute(sub-task)}
    \end{minipage}%
}
\end{codebox}
\vspace{-1.5mm}
\begin{table}[htbp]
\vspace{2mm}
\fontsize{8pt}{10pt}
\centering
\caption{Executability, Feasibility, and Success Rates of LLM-based Autonomy and Human-Robot Collaboration}
\vspace{-1.5mm}
\label{tab:success_rates}
\begin{tabular}{@{}l@{\hspace{0mm}}c@{\hspace{2mm}}c@{\hspace{2mm}}c@{\hspace{2mm}}c@{\hspace{2mm}}}
\toprule

Tasks & Num of trials & Executability & Feasibility & Success rate\\
\midrule
Put\&Stack & 23 & 100.0\% & 100.0\% & 91.3\%  \\
Open microwave & 23 & 100.0\% & 100.0\% & 82.6\%  \\
Open oven (HRC) & 23 & 100.0\% & 100.0\% & 91.3\% \\
Open cabinet (HRC) & 23 & 100.0\% & 100.0\% & 87.0\%  \\
\midrule
Clean table & 23 & 100.0\% & 95.7\% & 87.0\%  \\
Warm up apple & 23 & 100.0\% & 100.0\% & 60.9\%  \\
Roast apple (HRC) & 23 & 95.7\% & 87.0\% & 56.5\%  \\
\midrule
Total & 161 & 99.4\% & 97.5\% & 79.5\% \\
\bottomrule
\end{tabular}
\vspace{-2mm}
\end{table}
\vspace{-2mm}
\begin{table}[htbp]
\vspace{2mm}
\fontsize{8pt}{10pt}
\centering
\caption{Comparison of Executability, Feasibility, and Success Rates for Sub-Tasks With and Without HRC}
\vspace{-1.5mm}
\label{tab:comparison}
\begin{tabular}{@{}l@{\hspace{0mm}}c@{\hspace{2mm}}c@{\hspace{2mm}}c@{\hspace{2mm}}c@{\hspace{2mm}}}
\toprule
Tasks & Num of trials & Executability & Feasibility & Success rate\\
\midrule
Open oven & 1 & 100.0\% & 0.0\% & 0.0\% \\
Open oven (HRC) & 23 & 100.0\% & 100.0\% & 91.3\% \\
\midrule
Open cabinet & 1 & 100.0\% & 0.0\% & 0.0\% \\
Open cabinet (HRC) & 23 & 100.0\% & 100.0\% & 87.0\% \\
\bottomrule
\end{tabular}
\vspace{-3mm}
\end{table}
\subsection{Discussion}
All of the zero-shot basic tasks and the one-shot DMP-based tasks are indicated in Table 1 to evaluate their performance. A total of seven distinct tasks were assessed, with each task undergoing 23 trials, totaling 161 trials.

The \textbf{``Executability" }score reflects whether the integrated Pythonic code follows the predefined format and is executable. This score is quantitatively measured as the ratio of tasks for which the code is successfully executed to the total number of tasks assessed.

This score almost achieved 100\%, the only exception was noted in the task ``Roast apple (HRC)", where executability slightly decreased to 95.7\%. This was attributed to the LLM's occasional generation of sub-tasks without assigning corresponding motion functions, leading to non-encodable responses. 
However, this issue was not observed in subsequent experiments, suggesting its rarity and isolating it as a singular incident within the scope of our testing (one failure in 161 experiments). This indicates that the instance of reduced executability for ``Roast Apple (HRC)" was an anomaly rather than an indication of a systemic flaw.

The \textbf{``Feasibility"} score addresses whether the motion functions arranged by the LLM could accomplish the tasks under the assumption that there are no motion errors in robot control and no spatial errors in object recognition. To quantify this, this score is calculated as the percentage of tasks that meet the established standard out of the total number of tasks attempted.

This score highlights the indispensable role of HRC, as evidenced in Table \ref{tab:comparison}. Specifically, tasks like ``Open oven" and ``Open cabinet," initially deemed infeasible with a Feasibility of 0\% (impossible to be finished with motion functions from the basic library), witnessed remarkable improvements upon integrating HRC, underscoring HRC's enhancement on task execution.

Additionally, certain tasks such as ``Clean table" and ``Roast apple (HRC)" scored 95.7\% and 87.0\%, respectively. The ``Clean table" task's diminished score can be attributed to the YOLO's occasional oversight of objects during environmental scanning, which results in passing the incomplete environment information to the LLM. For ``Roast apple (HRC)", the issue arose from incorrect DMP function calls, exemplified by the LLM mistakenly activating \textquotesingle dmp\_pub(close\_oven\_handle)\textquotesingle\ instead of \textquotesingle dmp\_pub(close\_oven)\textquotesingle. Such inaccuracies, driven by token variability, can dilute the precision of motions reproduced from the DMP library, particularly in complex tasks prone to synonymous token substitution.

The \textbf{``Success rate''} score, which indicates the percentage of tasks completed in the real world, varied depending on the complexity of the task.
To investigate the cause of the reduction in tasks ``Warm up apple'' and ``Roast apple (HRC)'', experiments to test the performance of YOLO were conducted. In these experiments, the center of YOLO's bounding box for an object was utilized to represent its position. Additionally, AR markers were placed at the absolute center of the object's surface facing the robot as a ground truth position. Discrepancies between the object positions detected by YOLO and the ground truth positions indicated by the AR markers were measured over 5 seconds.

For specific tasks, such as placing a 0.063m high apple into a 0.086m high oven with a margin of 0.0115m and a discrepancy range of 0.0095m to 0.0122m, and a median discrepancy of 0.01m to 0.012m, the margin falls within the discrepancy range. This issue also occurs in other tasks like rotating the knob, opening the microwave door fully, and stacking the cups, which can affect the success rate of task execution. The errors of environment perception stem from YOLOv5's bounding box inaccuracies, leading to slightly variable coordinates for target objects and occasionally exceeding the necessary margins for precise manipulation.
Furthermore, even though the success rates of all sub-tasks in our experiments are above 80\%, the success rates of the long-horizon tasks will inevitably decrease due to error accumulation.
\vspace{-2mm}

\section{Conclusion}
This research advances robot autonomy through an interface linking the LLM with robotic systems, enabling task execution through natural language. Moreover, a novel LLM-based hierarchical planning framework efficiently manages long-horizon tasks, showcasing advanced task planning.
Furthermore, the incorporation of teleoperation and DMP into the HRC framework enables the enhancement of LLM-based robotic manipulation through human demonstrations.

Experimental results show the effectiveness of the proposed method: an average success rate of 79.5\%, with 99.4\% executability and 97.5\% feasibility across various tasks. 
These results underscore the system's robustness in translating language commands into robot motions and integrating operator instructions to accomplish unachievable tasks, making significant strides toward improving the performance of the LLM-based robot with the complexities of real-world task demands.
However, to address a limitation of current LLM-based robots that rely heavily on visual inputs, future research will focus on integrating LIDAR-derived point clouds and tactile sensing technologies to enhance the proposed LLM-based robot performance in real-world environments.
\vspace{-2mm}
\bibliographystyle{IEEEtran}
\bibliography{citation.bib}

\end{document}